%% file: effective_dimensionality.tex
\PassOptionsToPackage{unicode}{hyperref}
\PassOptionsToPackage{hyphens}{url}
\PassOptionsToPackage{dvipsnames,svgnames,x11names}{xcolor}
\documentclass[
  aps,
  pre,
  reprint,
  superscriptaddress,
  amsmath,
  amssymb,
  floatfix,
  longbibliography]{revtex4-2}
\usepackage{xcolor}
\usepackage{amsmath,amssymb}
\usepackage{iftex}
\ifPDFTeX
  \usepackage[T1]{fontenc}
  \usepackage[utf8]{inputenc}
  \usepackage{textcomp} % provide euro and other symbols
\else % if luatex or xetex
  \usepackage{unicode-math} % this also loads fontspec
  \defaultfontfeatures{Scale=MatchLowercase}
  \defaultfontfeatures[\rmfamily]{Ligatures=TeX,Scale=1}
\fi
\usepackage{lmodern}
\ifPDFTeX\else
\fi
\IfFileExists{upquote.sty}{\usepackage{upquote}}{}
\IfFileExists{microtype.sty}{% use microtype if available
  \usepackage[]{microtype}
  \UseMicrotypeSet[protrusion]{basicmath} % disable protrusion for tt fonts
}{}
\usepackage{graphicx}
\makeatletter
\newsavebox\pandoc@box
\newcommand*\pandocbounded[1]{% scales image to fit in text height/width
  \sbox\pandoc@box{#1}%
  \Gscale@div\@tempa{\textheight}{\dimexpr\ht\pandoc@box+\dp\pandoc@box\relax}%
  \Gscale@div\@tempb{\linewidth}{\wd\pandoc@box}%
  \ifdim\@tempb\p@<\@tempa\p@\let\@tempa\@tempb\fi% select the smaller of both
  \ifdim\@tempa\p@<\p@\scalebox{\@tempa}{\usebox\pandoc@box}%
  \else\usebox{\pandoc@box}%
  \fi%
}
\def\fps@figure{htbp}
\makeatother
\makeatletter
\@ifpackageloaded{caption}{}{\usepackage{caption}}
\AtBeginDocument{%
\ifdefined\contentsname
  \renewcommand*\contentsname{Table of contents}
\else
  \newcommand\contentsname{Table of contents}
\fi
\ifdefined\listfigurename
  \renewcommand*\listfigurename{List of Figures}
\else
  \newcommand\listfigurename{List of Figures}
\fi
\ifdefined\listtablename
  \renewcommand*\listtablename{List of Tables}
\else
  \newcommand\listtablename{List of Tables}
\fi
\ifdefined\figurename
  \renewcommand*\figurename{Figure}
\else
  \newcommand\figurename{Figure}
\fi
\ifdefined\tablename
  \renewcommand*\tablename{Table}
\else
  \newcommand\tablename{Table}
\fi
}
\@ifpackageloaded{float}{}{\usepackage{float}}
\floatstyle{ruled}
\@ifundefined{c@chapter}{\newfloat{codelisting}{h}{lop}}{\newfloat{codelisting}{h}{lop}[chapter]}
\floatname{codelisting}{Listing}

\makeatother
\makeatletter
\@ifpackageloaded{caption}{}{\usepackage{caption}}
\@ifpackageloaded{subcaption}{}{\usepackage{subcaption}}
\makeatother
\usepackage{bookmark}
\IfFileExists{xurl.sty}{\usepackage{xurl}}{} % add URL line breaks if available
\hypersetup{
  pdftitle={-VAEs as Effective Theories: Tolerance-Dependent Dimension},
  pdfauthor={Johannes Hirn},
  colorlinks=true,
  linkcolor={blue},
  filecolor={Maroon},
  citecolor={Blue},
  urlcolor={Blue},
  pdfcreator={LaTeX via pandoc}}

\begin{document}

\title{\texorpdfstring{$\beta$}{beta}-VAEs as Effective Theories:
Tolerance-Dependent Dimension}

\author{Johannes Hirn}

\affiliation{Image Processing Laboratory (IPL), Universitat de València,
Paterna, València 46980, Spain}

\email{hirnjo@uv.es}

\homepage{https://orcid.org/0000-0003-0267-2479}

\begin{abstract}
In a \(\beta\)-VAE, increasing the regularization strength acts as a
spectral cutoff by collapsing low-utility latent coordinates.

In the linear Gaussian VAE, the collapse order matches the ranking of
reconstruction utilities exactly, because both are set by the PCA
spectrum. We ask which parts of this picture survive in fully connected
nonlinear VAEs trained on WorldClim.

We find that nonlinear interactions shift and broaden collapse onsets,
so thresholds no longer coincide exactly with utilities. However, the
common ordering is preserved over the resolved ranks, so the spectral
cutoff still acts as a utility cutoff and the effective-description
logic carries through.

The resulting effective-dimension curves reveal a head--tail tradeoff:
increasing depth concentrates utility into the first few coordinates but
worsens tail fidelity.
\end{abstract}

\maketitle

\providecommand{\UtilityThresholdRsq}{0.98}
\providecommand{\UtilityThresholdRsqFull}{0.9800}
\providecommand{\UtilityThresholdRsqRanks}{0}
\providecommand{\UtilityThresholdRsqRankMin}{1}
\providecommand{\UtilityThresholdRsqRankMax}{0}
\IfFileExists{tables/utility_threshold_r2_metrics.tex}{\input{tables/utility_threshold_r2_metrics.tex}}{}

\section{Introduction}\label{introduction}

Many high-dimensional datasets admit descriptions that are substantially
simpler than their ambient representation
\citep{tenenbaum2000global, roweis2000nonlinear, cayton2005algorithms, bengio2013representation}.
The manifold hypothesis summarizes this observation by saying that data
often lie near lower-dimensional structure. The word ``near'' matters:
it leaves open how much residual variation is treated as noise rather
than modeled explicitly. In the idealized sharp-gap limit, a small
number of coordinates captures the dataset, while additional coordinates
contribute only near the noise floor.

In real datasets, however, there may be no sharp gap. In such cases,
effective dimension is not a single integer, but depends on the task,
model class, and tolerated error.

A tolerance is only meaningful once a task and distortion metric have
been fixed. When no downstream objective is specified, reconstruction is
the natural baseline task: it measures how much input variation is
retained, even though it can preserve redundant or nuisance variation
because these remain part of the target.

Here we fix reconstruction as the task, reconstruction distortion as the
metric, and fully connected \(\beta\)-VAEs as the model class, with
depth as the main architectural variable.

With task, metric, and model class fixed, tolerance becomes the main
variable. We ask how effective dimension changes as the allowed
reconstruction error is varied. This suggests an effective-theory
viewpoint: choosing a reconstruction tolerance amounts to choosing a
resolution, so that some variation is represented explicitly while the
rest is left unresolved in the residual term. Unlike in field theory
however, the residual is not absorbed into counterterms to keep
predictions cutoff-independent; it is allowed to increase as the cutoff
is raised.

In that broad Wilsonian sense, this is the effective-theory logic we use
here: the variables needed in the description can change with the chosen
cutoff \citep{wilson1974renormalization, goldenfeld2018lectures}. In a
VAE, the KL weight \(\beta\) plays the role of a tunable cutoff over
latent coordinates. This is a cutoff language for finite latent degrees
of freedom, not a full field-theory framework of counterterms and
cutoff-independent observables. It also differs from RG analogies for
deep networks, where layers are compared to coarse-graining
transformations \citep{mehta2014exact, lin2017why}: here the question is
not layer-by-layer flow, but the cutoff-dependent survival of latent
coordinates.

The measured object is therefore the rank--distortion curve: reading it
at fixed tolerance \(\epsilon\) gives the effective dimension
\(d_\mathrm{eff}(\epsilon)\), while the associated marginal utilities
give the ordered spectrum of reconstruction importance.

But a rank--distortion curve by itself is not enough. For the
corresponding coordinates to serve as effective variables, they need to
satisfy several properties: they should be shared across the dataset,
few enough to inspect, ordered by marginal contribution, and
controllable by a cutoff so that a chosen tolerance selects an active
subset. Sharp on/off thresholds are ideal but not required. Semantic
interpretation is a further step: the spectrum tells us which variables
deserve inspection, but it does not by itself name them
\citep{locatello2019challengingdisentanglement}.

Variational autoencoders (VAEs \citep{kingma2014autoencoding}) are
natural candidates here because their latent variables are probabilistic
degrees of freedom: the ELBO averages reconstruction over the
approximate posterior and charges a KL cost for keeping that posterior
input-dependent and distinct from the prior. At a chosen information
price, coordinates whose reconstruction benefit does not justify this
cost collapse toward the prior. This latent sparsity reduces an
overcomplete embedding to an active rank, which is read of as the
effective dimension at the corresponding reconstruction tolerance.

Collapse can also be caused or amplified by optimization effects or
decoder mismatch and is therefore often presented as a failure mode
\citep{dai2020usual, lucas2019dontblame}. Here we focus on the
equilibrium, \(\beta\)-controlled part of collapse and use it as a
selection mechanism. In a \(\beta\)-VAE
\citep{higgins2017betavae, burgess2018understandingbetavae}, \(\beta\)
is the price, in reconstruction-loss units, of one nat of latent
information. After normalization by the data variance in
decoder-variance units, \(T\) is the same price in normalized
reconstruction units.

The linear Gaussian analysis in Ref. \citep{hirn2026posteriorcollapse}
shows that this selection picture is exact in the linear VAE case. As
\(T\) is raised, latent coordinates collapse one by one. Each collapse
threshold equals both the marginal reconstruction utility of that
coordinate and the corresponding PCA explained-variance ratio. The
collapse spectrum, PCA spectrum, and pruning-utility spectrum therefore
coincide component by component.

This paper asks which parts of this exactly calibrated picture survive
in nonlinear VAEs. We test this on the WorldClim bioclimatic dataset
\citep{fick2017worldclim2}, using rank--distortion curves to ask how
many variables are needed to reach a given normalized reconstruction
error. Dinnage studies the same dataset with a VAE at a fixed operating
point and asks how useful the learned latent representation is for
downstream analysis \citep{dinnage2023worldclim}. That is a natural
question once one commits to a single bottleneck strength. Our question
is different: we vary the regularization strength and ask how the active
rank and reconstruction-utility spectrum change with the cutoff.

To answer this question, we scan the regularization strength \(\beta\),
rank the latent coordinates, and measure how reconstruction utility and
active rank evolve with the cutoff.

\section{Nonlinear Latent
Spectroscopy}\label{nonlinear-latent-spectroscopy}

The nonlinear scan uses the same normalized control parameter as the
linear theory of Ref. \citep{hirn2026posteriorcollapse}. The VAE
objective is

\[
\mathcal L = D+\beta R,
\]

where \(D\) is reconstruction distortion, i.e.~the reconstruction
negative log-likelihood, and \(R\) is the latent KL, or information
rate. The raw parameter \(\beta\) sets the strength of the KL
regularization. An overbar denotes an average over data samples, and
brackets denote posterior averages.

\[
D=-\overline{\left\langle \log p_\theta(x\mid z)\right\rangle},
\qquad
R=\overline{\mathrm{KL}\!\left[q_\phi(z\mid x)\,\|\,p(z)\right]}.
\]

For the fixed-variance Gaussian decoder used here,

\[
D=
\frac{1}{2\sigma_\mathrm{dec}^2}
\overline{\left\langle
\|x-\hat x_\theta(z)\|^2
\right\rangle}
\;+\;\text{const},
\]

To compare \(\beta\) meaningfully with the data, and with the linear
baseline of Ref. \citep{hirn2026posteriorcollapse}, it must be measured
against the total data variance in decoder-variance units,
\(V/\sigma_\mathrm{dec}^2\). For a fixed-variance Gaussian decoder, we
therefore use the normalized control parameter

\[
T
=
\frac{\beta}{V/\sigma_\mathrm{dec}^2}
=
\frac{\beta\sigma_\mathrm{dec}^2}{V},
\]

where \(V\) is the total data variance. Thus \(T\) is the information
price in normalized distortion units.

The KL regularization acts spectrally: coordinates collapse one by one
as this price is raised, rather than being uniformly shrunk. In the
linear case this cutoff is exactly calibrated by utility; in the
nonlinear case this calibration is part of what we test below.

The thermodynamic analogy has a precise but limited scope here. The
latent variables are internal probabilistic degrees of freedom averaged
over in the ELBO through the expectation over \(q_\phi(z\mid x)\) and
through the KL term relative to the prior. By contrast, the encoder and
decoder parameters are not averaged over: they are selected by
optimization of the objective.

This means that the VAE is not a thermodynamic or field-theoretic system
in the usual sense, and thus has no large-\(N\) or continuum limit. The
useful analogy is instead one of effective description with a finite
number of learned internal variables: some variation is represented
explicitly through active latent coordinates, while the rest is absorbed
into the residual distortion.

In the linear Gaussian case this structure becomes exactly solvable mode
by mode; in the nonlinear case it remains a useful language for thinking
about cutoff, activity, and ranked utility without becoming a literal
statistical-mechanics construction.

For each latent coordinate \(k\), we monitor the posterior mean-square
amplitude, posterior variance, posterior log-variance, and rate,

\[
\overline{\mu_k(x)^2},\qquad
\overline{\sigma_k(x)^2},\qquad
\overline{\log\sigma_k(x)^2},\qquad
R_k(T).
\]

The primary order parameter is the scale-invariant signal fraction

\[
M_k^2(T)=
\frac{\overline{\mu_k(x)^2}}
{\overline{\mu_k(x)^2}+\overline{\sigma_k(x)^2}}.
\]

The denominator is the aggregate second moment
\(A_k^2=\overline{\langle z_k^2\rangle}=\overline{\mu_k(x)^2+\sigma_k(x)^2}\),
so \(M_k^2\) measures the signal fraction of the aggregate coordinate
without assuming a fixed latent scale.

The general idea is to call a coordinate active once an order parameter
shows a clear departure from the collapsed prior-like branch. In the
scan diagnostics used here, we classify coordinate \(k\) as active when
\(M_k^2>0.1\). This choice is not unique: one could instead threshold
the SNR-like score, the rate, or the posterior log-variance. We checked
that such choices shift exact thresholds and active counts slightly, but
do not change the qualitative ranking or the concentration of utility in
the leading coordinates.

Equivalently,

\[
M_k^2(T)=\frac{\mathrm{SNR}_k(T)}{1+\mathrm{SNR}_k(T)},
\qquad
\mathrm{SNR}_k(T)=
\frac{\overline{\mu_k(x)^2}}{\overline{\sigma_k(x)^2}}.
\]

This ratio is invariant under a common rescaling of the posterior mean
and standard deviation of a latent coordinate, so it provides a
scale-free order parameter for collapse.

For squared-error reconstruction, we report normalized distortion in the
same convention as Ref. \citep{hirn2026posteriorcollapse}, \[
\tilde D_K \equiv \frac{\overline{\|x-\hat x^{(K)}\|^2}}{V},
\] where \(\hat x^{(K)}\) is the reconstruction obtained after retaining
the first \(K\) ranked latent coordinates and pruning the rest. For a
fixed ranking, these truncated reconstructions define the nonlinear
utility spectrum. The marginal utility of rank \(K\) is

\[
\Delta\tilde D_K=\tilde D_{K-1}-\tilde D_K.
\]

The ranked sequence \(\{\Delta\tilde D_K\}\) is the spectrum compared
across architectures. It is empirical: it depends on the trained
nonlinear representation, the scan protocol, and the chosen ranking
observable.

\section{Linear Baseline and Nonlinear Spectral
Deformation}\label{linear-baseline-and-nonlinear-spectral-deformation}

The linear Gaussian VAE provides the calibrated baseline. The
reconstruction geometry is quadratic and diagonal in the PCA basis, so
each eigendirection behaves independently, and activating one mode thus
does not change the utility of another. The signal-fraction branch
follows the one-mode form

\[
M_k^2(T)=\left[1-\frac{T}{T_k}\right]_+.
\]

and the collapse threshold, reconstruction utility, and PCA weight
coincide:

\[
T_k=\Delta\tilde D_k=\frac{\lambda_k}{V}.
\]

The linear result is a calibrated null model. Assuming a centered
dataset, a fixed-variance Gaussian decoder, a standard normal prior, and
a linear encoder--decoder pair, the equilibrium \(\beta\) scan does not
merely recover the PCA subspace. It resolves the PCA spectrum mode by
mode: the \(k\)-th latent becomes active for \(T_k \leq \lambda_k/V\),
its signal fraction follows \(M_k^2=[1-T/T_k]_+\), and retaining that
mode lowers the normalized distortion by the same amount,
\(\Delta\tilde D_k=T_k\). The nonlinear experiments below keep this
measurement protocol and ask which parts of this calibration remain once
the decoder can mix modes.

We call such departures nonlinear spectral deformations. Several
deformations are possible: collapse thresholds can shift relative to
PCA, the coordinate order can change, leading modes can absorb a larger
fraction of the total distortion reduction, and onset branches can
soften into crossovers when a new coordinate activates in the background
created by already-active coordinates.

This last effect requires separating a local instability from a finite
reconstruction utility. Let \(A\) denote the set of already-active
coordinates, and let \(q_k\) be a small activity amplitude for a
collapsed coordinate \(k\). Locally, the objective can be expanded as

\[
\mathcal L_{A+k}
=
\mathcal L_A
+\frac{1}{2}\left[T-g_k(A,T)\right]q_k^2
+O(q_k^4),
\]

where \(g_k(A,T)\) is the local reconstruction gain available to
coordinate \(k\) in the background of the active set. The infinitesimal
onset is controlled by

\[
T=g_k(A,T).
\]

The finite utility measured by truncation is instead

\[
\Delta\tilde D_k(A)=\tilde D(A)-\tilde D(A\cup\{k\}).
\]

In the linear Gaussian case these two notions coincide because the
quadratic reconstruction operator is global and diagonal in the PCA
basis:

\[
g_k(A,T)=\Delta\tilde D_k(A)=\Delta\tilde D_k.
\]

The onset condition is therefore independent of which other modes are
active. This does not mean that earlier modes have saturated when later
modes appear: even in the linear branch, an already-active coordinate
continues to grow as \(T\) decreases. The special feature of the linear
case is that the local quadratic operator is fixed once and for all, so
the growth of one mode does not change the instability scale of another.

A nonlinear representation has no reason to obey this additivity. The
relevant quadratic operator is local to the current active background
and can change along the scan. Schematically,

\[
g_k(A,T)
=
g_k^{(0)}
+\sum_{\ell\in A}J_{k\ell}(T)
+\sum_{\ell,m\in A}J_{k\ell m}(T)
+\cdots ,
\]

and the finite utility need not equal this local gain at every point
along the branch. The coefficients in this expansion summarize how the
marginal usefulness of one coordinate is dressed by the nonlinear
reconstruction background built from the coordinates that are already
active. This explains why nonlinear onsets need not follow the linear
branch exactly. We now turn to the measured scans.

\section{Ranked Utility Spectrum in
WorldClim}\label{ranked-utility-spectrum-in-worldclim}

Our main empirical object is the ranked utility spectrum for WorldClim.
Unless noted otherwise, the single-architecture scans below use a
representative fully connected VAE with four hidden layers of width 64
in both encoder and decoder, and 32 latent coordinates. The depth
comparison keeps the width fixed at 64 and the latent dimension fixed at
32 while varying depth over \(2,4,8,\) and \(16\) hidden layers, so the
comparison isolates nonlinear composition rather than a broad
architecture sweep.

\begin{figure}

\centering{

\pandocbounded{\includegraphics[keepaspectratio]{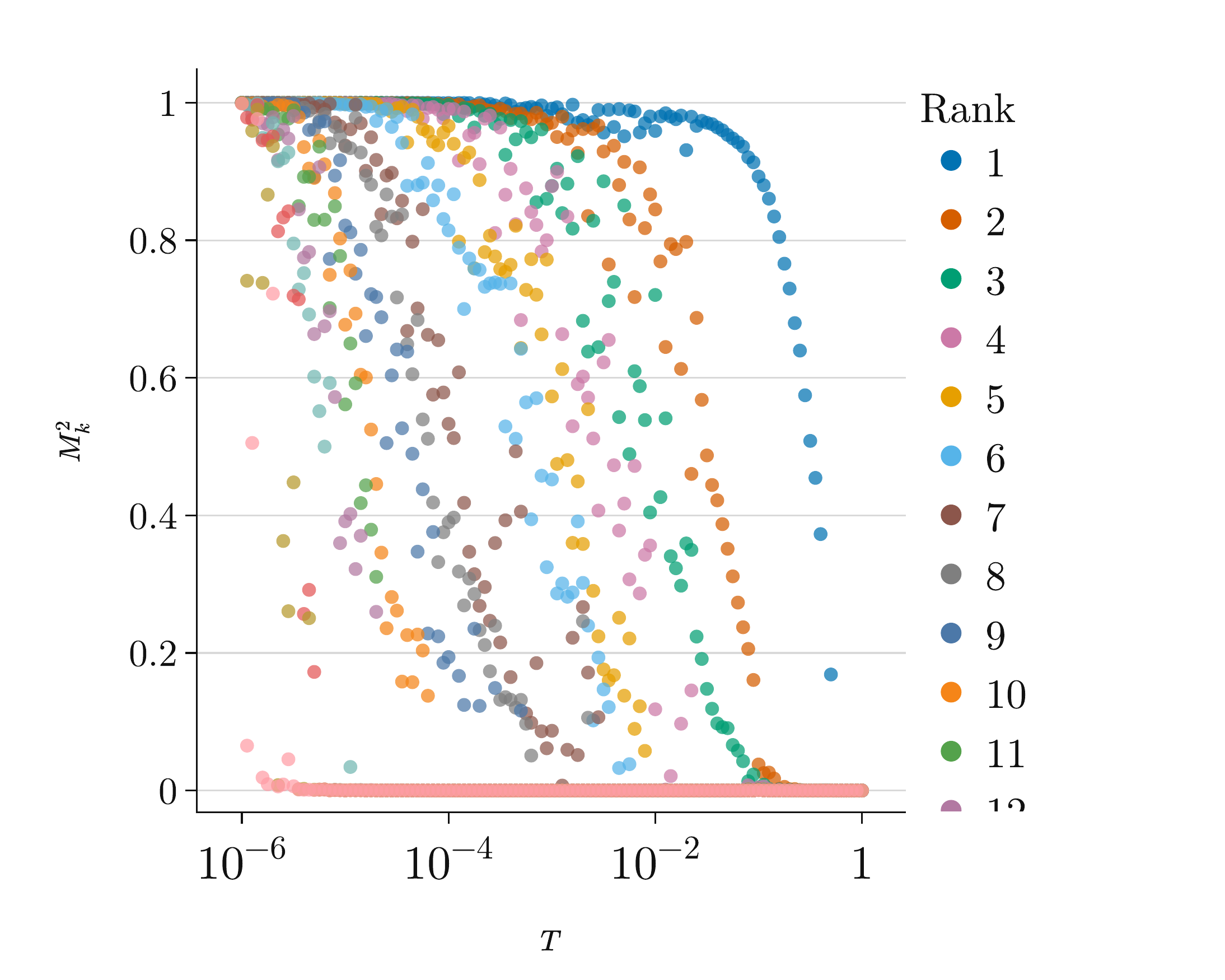}}

}

\caption{\label{fig-nonlinear-order-parameters}Signal-fraction scan for
the representative 4-layer VAE. Colors denote rank after sorting by the
SNR-like score, equivalently by final \(M_k^2\).}

\end{figure}

Fig. \ref{fig-nonlinear-order-parameters} shows the signal-fraction scan
used to rank coordinates. The first coordinates activate in a clear
sequence as \(T\) is lowered, but the branches do not appear to follow
exact one-mode Landau curves. In the nonlinear model, the onset can be
rounded or shifted by the active background and by optimization. We
therefore use \(M_k^2\) primarily as a scale-invariant ranking
observable, not as the main source of fitted thresholds.

\begin{figure}

\centering{

\pandocbounded{\includegraphics[keepaspectratio]{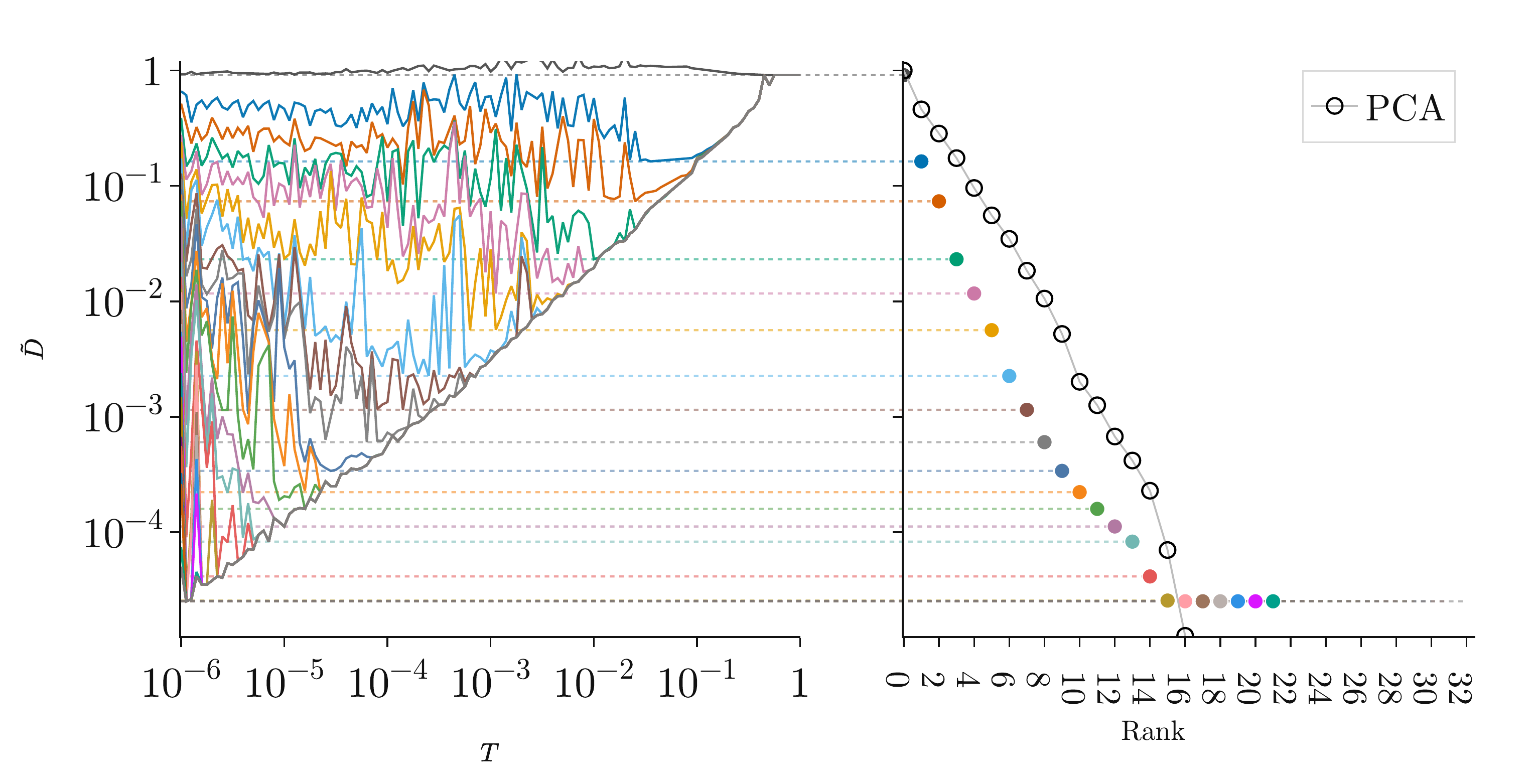}}

}

\caption{\label{fig-nonlinear-distortion-pruning}Truncated distortion
under ranked pruning. Left: distortion after retaining the first \(K\)
ranked latents at each temperature. Right: best distortion at each rank;
successive drops define \(\Delta\tilde D_K\).}

\end{figure}

Fig. \ref{fig-nonlinear-distortion-pruning} measures utilities directly.
For each point along the scan, we retain only the first \(K\) ranked
latents and prune the rest. The left panel follows the resulting
normalized distortion as a function of \(T\); the right panel records
the best distortion reached at each retained rank. The marginal drops in
this curve define \(\Delta\tilde D_K\). This step is independent of any
threshold fit.

This pruning measurement is a property of the trained representation,
not a comparison with separately retrained \(K\)-latent models. The
distinction is intentional: during training, reconstruction utility can
be distributed across more coordinates than are ultimately needed at a
chosen tolerance. The reported curve asks how much distortion remains
after truncating the trained representation to its first \(K\) ranked
variables.

Because the nonlinear onset shape is not known a priori, we do not treat
the rounded \(M_k^2\) branches as exact fit objects. Instead, we use
\(M_k^2\) to rank coordinates, measure utilities directly by pruning,
and test the linear active-branch form on the rate, where the scaling is
empirically cleanest. In the linear theory, the same number is both a
truncation utility and a collapse threshold. We therefore ask whether
the rate curves overlay when scaled by the measured utility:

\[
R_k(T)\simeq
\left[-\frac{1}{2}\log\frac{T}{\Delta\tilde D_k}\right]_+ .
\]

\begin{figure}

\centering{

\pandocbounded{\includegraphics[keepaspectratio]{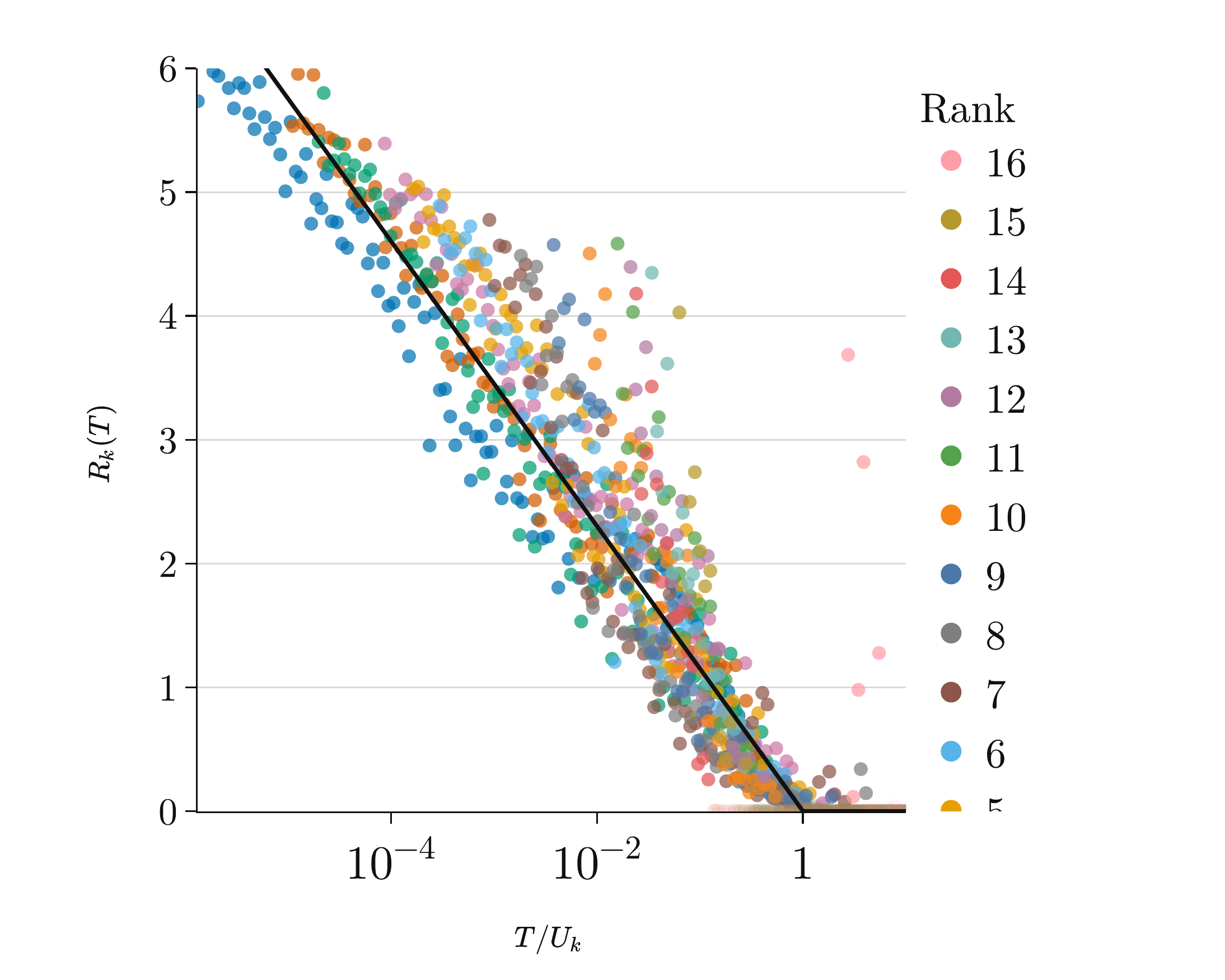}}

}

\caption{\label{fig-rate-utility-landau-overlay}Rate curves rescaled by
pruning utilities \(\Delta\tilde D_k\). The black curve is the one-mode
prediction; colored points are nonlinear branches ranked by \(M_k^2\).}

\end{figure}

Fig. \ref{fig-rate-utility-landau-overlay} shows this rescaled overlay
for the information rate \(R_k(T)\) using the utilities from Fig.
\ref{fig-nonlinear-distortion-pruning}. The black curve is the one-mode
Landau prediction with no fitted threshold. The colored points are still
ranked by \(M_k^2\), so the overlay also checks agreement between the
order-parameter ranking and the rate ranking. In diagnostic checks,
alternative rankings by information rate or posterior log-variance
mainly reorder weak or near-degenerate coordinates and do not change the
leading shortlist.

There is no parameter fitting involved in Fig.
\ref{fig-rate-utility-landau-overlay}: the point is that a utility
measured from reconstruction predicts the rate scale of the active
branch. At much lower relative temperature, roughly two decades below
threshold, some higher-rank branches steepen; this deep active tail is
not used for later fits of thresholds.

The same rescaled overlay identifies a natural active window for a
secondary threshold check. For each rank, we fit the unrescaled rate
branch to the fixed-slope form

\[
R_k(T)
\simeq
\left[-\frac{1}{2}\log\frac{T}{T_{c,k}^{\mathrm{rate}}}\right]_+
\]

over

\[
10^{-2}\Delta\tilde D_k
\leq
T
\leq
\Delta\tilde D_k .
\]

\begin{figure}

\centering{

\pandocbounded{\includegraphics[keepaspectratio]{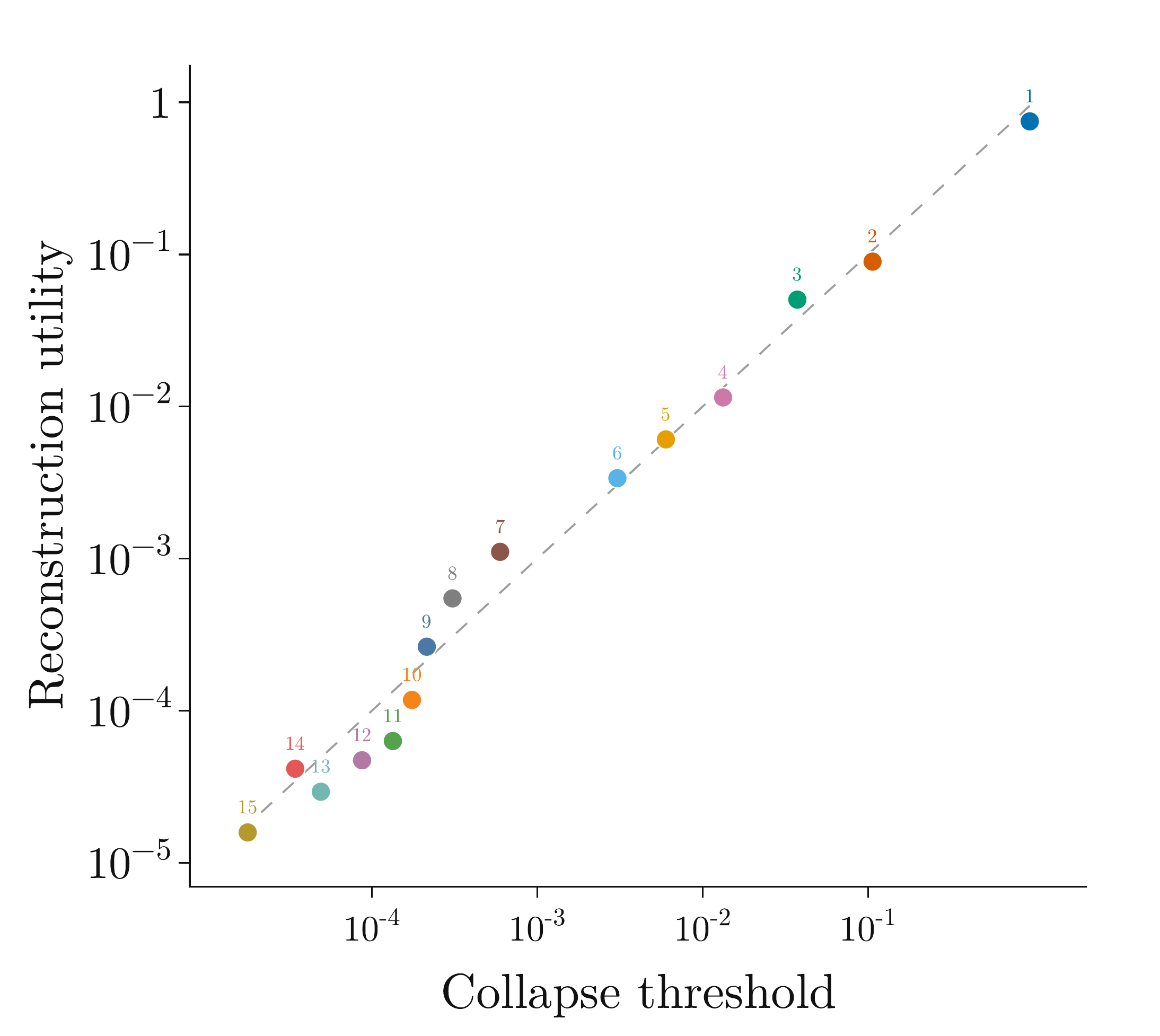}}

}

\caption{\label{fig-rate-utility-threshold}Utility--threshold comparison
from fixed-slope rate fits. Each point compares \(\Delta\tilde D_k\)
with the corresponding rate-fit threshold \(T_{c,k}^{\mathrm{rate}}\).}

\end{figure}

The window is defined from the pruning utility, but the intercept
\(T_{c,k}^{\mathrm{rate}}\) is extracted from the rate branch. Fig.
\ref{fig-rate-utility-threshold} compares these fitted thresholds with
the directly measured utilities. The diagonal agreement is summarized by
an identity \(R^2\) on log-scaled values, \(\log\Delta\tilde D_k\)
versus \(\log T_{c,k}^{\mathrm{rate}}\). This is a predictive identity
score on the log values, not the square of a correlation coefficient, so
overall offsets are penalized.

On the test split, the check gives \(R^2\simeq0.98{}\) over the plotted
ranks. For the representative WorldClim run, the threshold ranking
agrees with the utility ranking through the first thirteen modes; beyond
that point, neighboring fitted thresholds begin to swap while the
pruning utilities remain clearly separated. This is not the primary way
to define the spectrum, but a consistency check on the utility
calibration.

\begin{figure}

\centering{

\pandocbounded{\includegraphics[keepaspectratio]{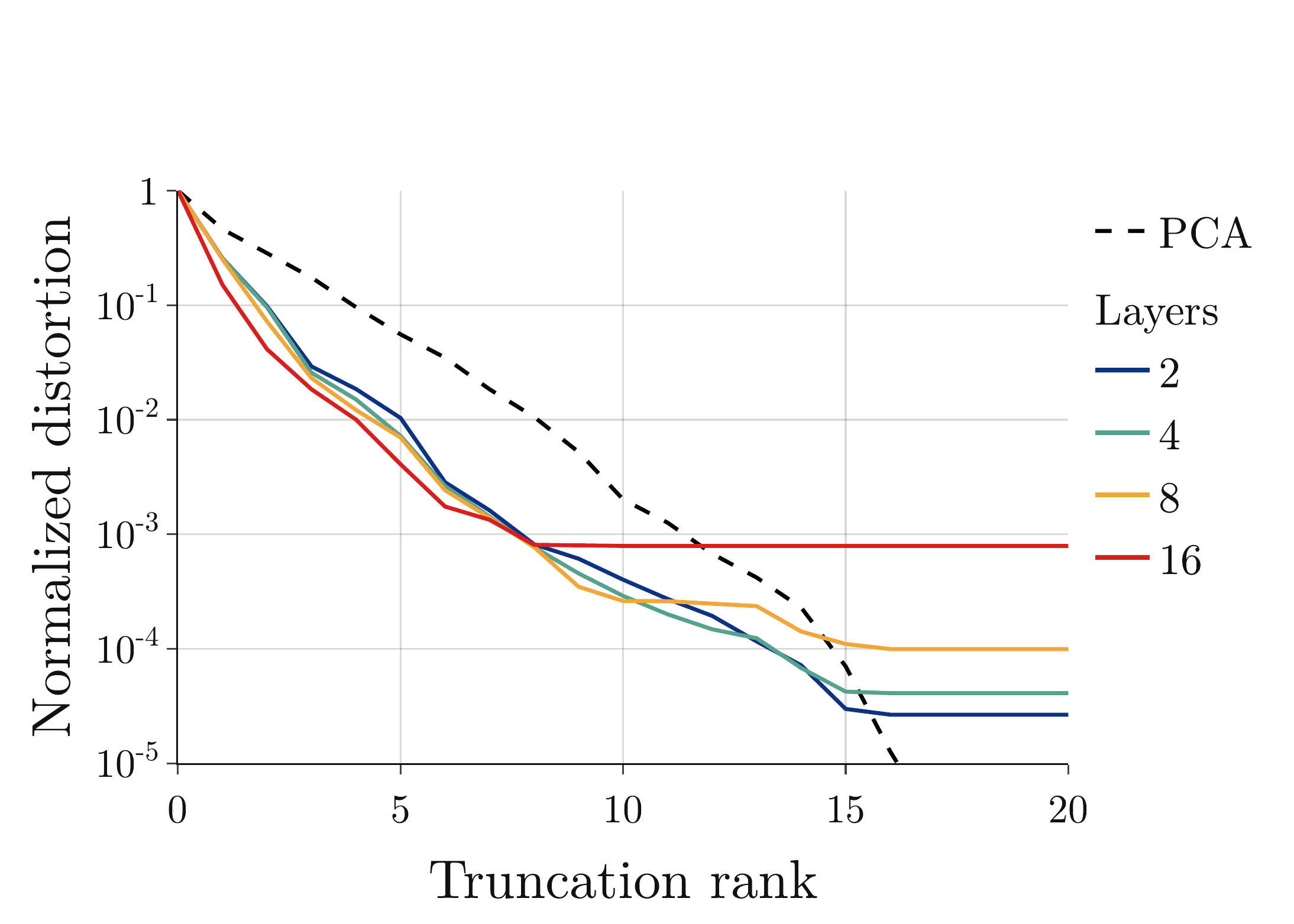}}

}

\caption{\label{fig-rank-normalized-distortion}Rank--distortion curves
under ranked pruning. Lower curves indicate a more concentrated utility
spectrum.}

\end{figure}

A compact sparsity diagnostic is the required rank at a prescribed
normalized distortion. Fig. \ref{fig-rank-normalized-distortion} shows
the measured rank--distortion curves. Lower curves are more spectrally
sparse: they recover the same distortion reduction with fewer retained
variables. This comparison is clearest on a linear rank axis and
logarithmic distortion axis, because it exposes both the low-rank head
and the residual tail. The plotted comparison fixes the hidden width and
varies depth. The first few ranks form a depth-sensitive head:
increasing depth can concentrate more utility into the leading effective
variables. Beyond rank 8, however, the deepest models often saturate at
a higher residual floor.

Reading the same curves at fixed tolerance gives the effective
dimension,

\[
d_\mathrm{eff}(\epsilon)=\min\{K: D(K)/D_0\le \epsilon\}.
\]

Table \ref{tbl-rank-precision-summary} reports this fixed-tolerance
read-off from \(5\%\) down to \(10^{-5}\).

\input{tables/rank_precision_summary.tex}

\begin{figure}

\centering{

\pandocbounded{\includegraphics[keepaspectratio]{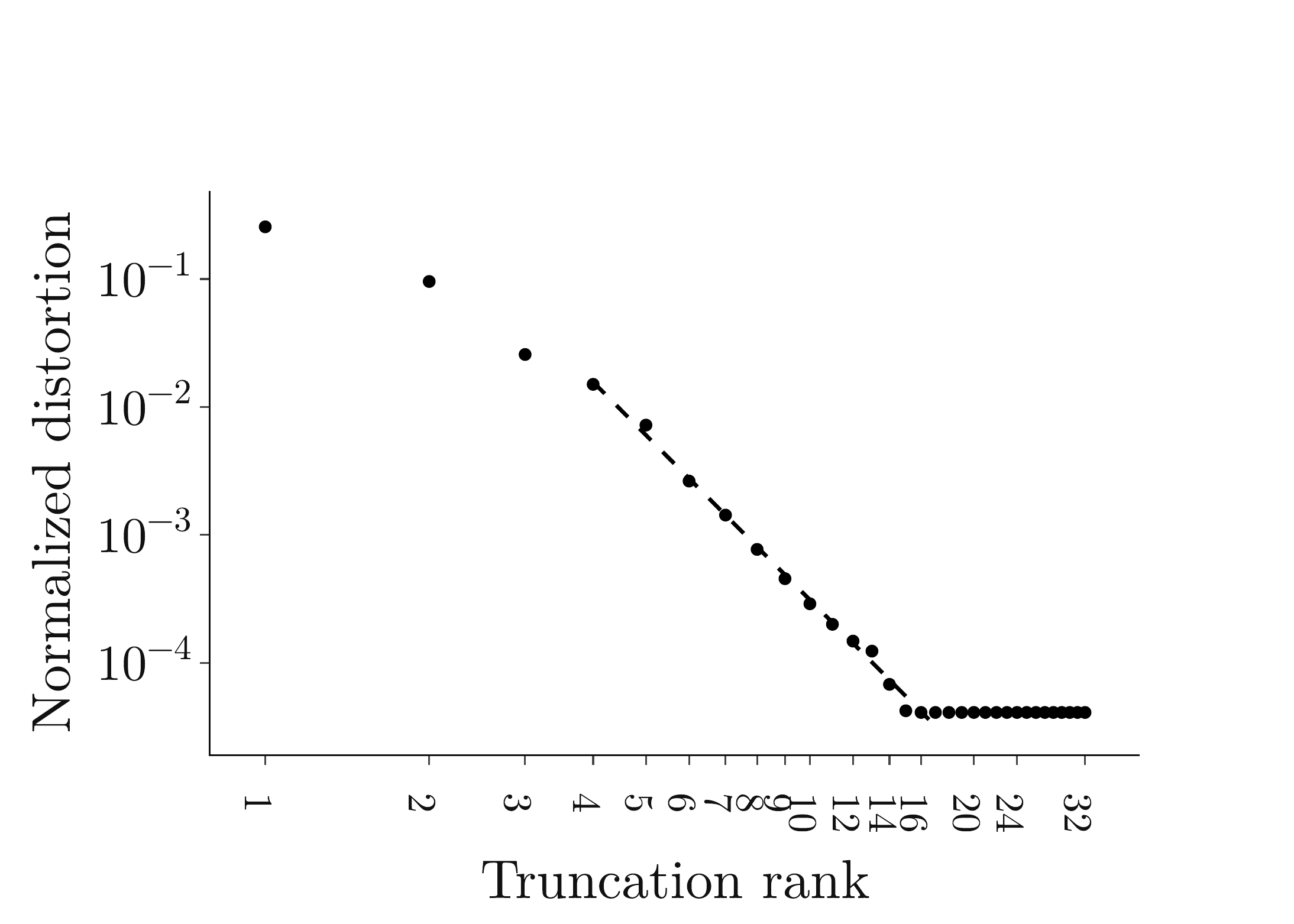}}

}

\caption{\label{fig-rank-distortion-power-law}Log--log rank--distortion
curve for the representative four-layer width-64, 32-latent VAE. The
dashed line is the power-law fit over the chosen rank window.}

\end{figure}

Fig. \ref{fig-rank-distortion-power-law} suggests that the same fall-off
can be described locally by a power law. On log--log axes, the
head--tail separation appears as a crossover near \(K_c\simeq4\): the
first few ranks form the head, and subsequent ranks follow the residual
tail until architecture-dependent floors appear. We interpret \(K_c\)
only as a spectral crossover rank, not as a unique intrinsic dimension
of the dataset. Fig. \ref{fig-rank-distortion-power-law} shows the
representative four-layer width-64, 32-latent VAE on log--log axes, with
a fit over the low-rank window used in the architecture comparison. For
WorldClim, the residual distortion falls approximately as
\(D(K)\sim K^{-4}\) over ranks \(4\)--\(17\). The corresponding spectral
exponent is closer to \(5\) since the marginal utility spectrum
\(\Delta\tilde D_K\) is the discrete derivative of \(D(K)\) and is
therefore roughly one power steeper.

For this protocol, the full threshold spectrum is not required. The
direct object is the truncation curve: train across a grid of normalized
spectral cutoffs, rank the latent variables by a stable order parameter,
and record the normalized distortion reached after retaining the first
\(K\) variables. At fixed width, increasing depth mostly sharpens the
low-rank head of this curve: more utility is concentrated into the first
few retained variables. The same increase in depth is not uniformly
beneficial in the tail, since the deepest models can plateau at higher
residual distortion. We therefore describe depth as a head--tail
tradeoff at fixed width, not as a monotonic improvement.

\section{Conclusions and Outlook}\label{conclusions-and-outlook}

For WorldClim, the nonlinear scan confirms the cutoff picture, but it
also shows that direct reconstruction utilities are more robust than
fitted collapse thresholds. Utilities are therefore the preferred
spectrum: they directly measure the feature importance of each ranked
latent coordinate, remain usable at higher ranks, and avoid the
fit-window and outlier sensitivity of threshold extraction. Thresholds
serve mainly as a check of the utility--threshold duality.

This changes the practical role of the scan. The scan is the calibration
step, not the end goal. Once the spectral-cutoff picture has been
checked for a given model class and normalization, one need not extract
thresholds or run a full scan on every dataset. Instead, one can choose
the normalized regularization strength to match the smallest
reconstruction utility one is willing to trust, for example a noise
floor or modeling tolerance.

Under this interpretation, the main output is not a single preferred
bottleneck size, but a tolerance-dependent effective dimension.
Effective dimension is therefore not a fixed property of the data alone;
it depends on the task, distortion metric, model class, and tolerated
reconstruction error.

Architecture changes the shape of that dependence: at fixed width, depth
mainly redistributes utility toward the head of the spectrum, often at
the cost of a higher tail floor. This is a latent sparsity tradeoff:
deeper models concentrate more reconstruction value into the first few
coordinates, but can leave a less efficient residual tail.

These results support an effective-theory reading of the \(\beta\)-VAE
scan. The value of \(T\) specifies the price of resolving latent
information; the trained model then decides which learned variables
remain explicit and which variation is absorbed into the residual
distribution. In that sense, the scan does not assume an effective
description in advance; it reveals one by showing which variables
survive at the chosen cutoff.

This selection mechanism also clarifies what distinguishes VAEs from
deterministic and sparse autoencoders. VAEs introduce probabilistic
latent degrees of freedom, average over them in the objective, and
penalize their departure from a prior. This is structurally close to a
finite-dimensional variational free-energy problem over latent
variables, in a way that deterministic embeddings are not. A
deterministic autoencoder can learn a low-dimensional bottleneck, but
the bottleneck dimension is fixed by design: the model tests whether
that chosen dimension is sufficient, not which coordinates should
survive as the tolerance changes. Sparse autoencoders define a different
notion of sparsity that can be useful in other contexts
\citep{bricken2023sparse, cunningham2023sparse}: they impose sparsity at
the level of individual codes, with each input activating only a small
subset of dictionary elements, and the active subset changing from one
data point to the next. This is well suited to feature discovery,
especially in overcomplete representations, but it does not by itself
produce a small globally ranked set of coordinates shared across the
dataset. Other regularizers can also suppress dimensions or produce
sparse codes, but they generally do not retain the same prior-relative,
information-theoretic interpretation of the thresholds.

In the VAE, this cutoff picture comes from a prior-relative rate cost
for making an entire coordinate input-dependent across the dataset. A
coordinate either earns its KL cost or collapses toward the prior
globally. Rather than asking whether a system is emergent in the
abstract, one can ask how many effective variables are required to
reproduce the behavior of interest at a given tolerance.

Because the present work ranks variables by reconstruction utility, it
can retain redundant or nuisance variation whenever that variation helps
reproduce the input distribution. A natural next step is therefore to
embed the same variational bottleneck inside a supervised predictor and
rank variables by task utility. Since the bottleneck would then lie on
the prediction path, pruning would measure the importance of variables
for the model actually used for inference, rather than for a post-hoc
surrogate.

In that setting, choosing \(\beta\) becomes an operating-point question:
one would trade off task performance, active dimension, and the
interpretability of the surviving coordinates.

\section{Appendix: WorldClim Experiment
Details}\label{sec-worldclim-experiment-details}

The WorldClim experiments use the 19 bioclimatic variables from
WorldClim \citep{fick2017worldclim2} at 10 arc-minute resolution. We
keep grid cells that are valid in all 19 layers and standardize each
feature using the training split only. The data split uses seed 0 and
the spatial-block protocol used throughout these experiments: equal-area
blocks of side length 500 km, area-weighted sampling, a target of
500,000 training samples, and an equal split of the non-training blocks
between validation and test. The batch size is 256.

The representative nonlinear run used in Figs.
\ref{fig-nonlinear-order-parameters},
\ref{fig-nonlinear-distortion-pruning},
\ref{fig-rate-utility-landau-overlay}, and
\ref{fig-rate-utility-threshold} is the saved VAE run where the encoder
and decoder are fully connected networks with four hidden layers, 64
hidden units per layer, and 32 latent coordinates. We use SiLU
activations, Kaiming initialization, and initial log-variance bias
\(-4\). The reconstruction loss is mean-squared error with fixed decoder
variance convention \(\sigma_\mathrm{dec}^2=1\) and no additional
reconstruction normalization.

The scan evaluates an independent grid of normalized temperatures, with
\(T\) running from \(1\) to \(10^{-6}\). Each value of \(T\) is trained
from a fresh initialization rather than warm-started from neighboring
scan points, so the ordering of the grid is not an annealing path and
does not introduce scan hysteresis. Here ``equilibrium'' means that each
independently trained scan point is run until the configured convergence
criterion is met; it does not imply a proof of global optimality for the
nonlinear objective. In the saved resolved configuration this
corresponds to 121 scan points and raw \(\beta\) values from \(19.0\) to
\(1.9\times10^{-5}\), because the measured WorldClim variance scale is
\(V=19\). Distortions are reported in normalized squared-error units,
i.e.~reconstruction error divided by the total data variance \(V\). Each
scan point is trained with Adam at learning rate \(3\times10^{-4}\),
with a maximum budget of \(10^6\) optimizer updates, relative tolerance
\(10^{-3}\), and patience fraction \(0.01\). The reported quantities are
evaluated on the test split. PCA references use the same normalized
data.

Latent coordinates are ranked by the saved SNR-like score, equivalently
by the signal fraction \(M_k^2\), using descending score order. The
active threshold used for scan diagnostics is \(0.1\). Ranked-pruning
curves retain the first \(K\) coordinates in this order and prune the
rest.

\bibliography{../references.bib}

\end{document}

%% file: tables/utility_threshold_r2_metrics.tex
% Auto-generated by 05_analyze.py.
\renewcommand{\UtilityThresholdRsq}{0.98}
\renewcommand{\UtilityThresholdRsqFull}{0.9845}
\renewcommand{\UtilityThresholdRsqRanks}{15}
\renewcommand{\UtilityThresholdRsqRankMin}{1}
\renewcommand{\UtilityThresholdRsqRankMax}{15}

%% file: tables/rank_precision_summary.tex
\begin{table}[t]
\caption{Required truncation rank to reach each normalized distortion target. For each target, the VAE entry is the lowest rank among the compared VAE architectures. A dash indicates that the target is not reached.}
\label{tbl-rank-precision-summary}
\centering
\begin{tabular}{lcccccc}
\hline
 & \multicolumn{6}{c}{Distortion} \\
Model & 5\% & 1\% & 0.5\% & 0.1\% & 0.01\% & 0.001\% \\
\hline
PCA & 6 & 9 & 10 & 12 & 15 & 17 \\
VAE & 2 & 5 & 5 & 8 & 14 & \textemdash{} \\
\hline
\end{tabular}
\end{table}